Original Paper

# Leveraging Medical Knowledge Graphs Into Large Language Models for Diagnosis Prediction: Design and Application Study

Yanjun Gao[1,2], PhD; Ruizhe Li[3], PhD; Emma Croxford[2], BS; John Caskey[2], PhD; Brian W Patterson[2], MPH, MD; Matthew Churpek[2], MPH, MD, PhD; Timothy Miller[4], PhD; Dmitriy Dligach[5], PhD; Majid Afshar[2], MD, MSCR

[1]Department of Biomedical Informatics, University of Colorado Anschutz Medical Campus, Denver, CO, United States
[2]Department of Medicine, University of Wisconsin–Madison, Madison, WI, United States
[3]University of Aberdeen, Aberdeen, United Kingdom
[4]Boston Children's Hospital, Harvard Medical School, Boston, MA, United States
[5]Loyola University Chicago, Chicago, IL, United States

**Corresponding Author:**
Yanjun Gao, PhD
Department of Biomedical Informatics
University of Colorado Anschutz Medical Campus
1890 N Revere Ct
Denver, CO, 80045
United States
Phone: 1 303 724 5375
Email: yanjun.gao@cuanschutz.edu

## *Abstract*

**Background:** Electronic health records (EHRs) and routine documentation practices play a vital role in patients' daily care, providing a holistic record of health, diagnoses, and treatment. However, complex and verbose EHR narratives can overwhelm health care providers, increasing the risk of diagnostic inaccuracies. While large language models (LLMs) have showcased their potential in diverse language tasks, their application in health care must prioritize the minimization of diagnostic errors and the prevention of patient harm. Integrating knowledge graphs (KGs) into LLMs offers a promising approach because structured knowledge from KGs could enhance LLMs' diagnostic reasoning by providing contextually relevant medical information.

**Objective:** This study introduces DR.KNOWS (Diagnostic Reasoning Knowledge Graph System), a model that integrates Unified Medical Language System–based KGs with LLMs to improve diagnostic predictions from EHR data by retrieving contextually relevant paths aligned with patient-specific information.

**Methods:** DR.KNOWS combines a stack graph isomorphism network for node embedding with an attention-based path ranker to identify and rank knowledge paths relevant to a patient's clinical context. We evaluated DR.KNOWS on 2 real-world EHR datasets from different geographic locations, comparing its performance to baseline models, including QuickUMLS and standard LLMs (Text-to-Text Transfer Transformer and ChatGPT). To assess diagnostic reasoning quality, we designed and implemented a human evaluation framework grounded in clinical safety metrics.

**Results:** DR.KNOWS demonstrated notable improvements over baseline models, showing higher accuracy in extracting diagnostic concepts and enhanced diagnostic prediction metrics. Prompt-based fine-tuning of Text-to-Text Transfer Transformer with DR.KNOWS knowledge paths achieved the highest ROUGE-L (Recall-Oriented Understudy for Gisting Evaluation–Longest Common Subsequence) and concept unique identifier $F_1$-scores, highlighting the benefits of KG integration. Human evaluators found the diagnostic rationales of DR.KNOWS to be aligned strongly with correct clinical reasoning, indicating improved abstraction and reasoning. Recognized limitations include potential biases within the KG data, which we addressed by emphasizing case-specific path selection and proposing future bias-mitigation strategies.

**Conclusions:** DR.KNOWS offers a robust approach for enhancing diagnostic accuracy and reasoning by integrating structured KG knowledge into LLM-based clinical workflows. Although further work is required to address KG biases and extend generalizability, DR.KNOWS represents progress toward trustworthy artificial intelligence–driven clinical decision support, with a human evaluation framework focused on diagnostic safety and alignment with clinical standards.

(*JMIR AI 2025;4:e58670*)   doi: 10.2196/58670

XSL•FO
RenderX

## Introduction

### Background

The ubiquitous use of electronic health records (EHRs) and the standard documentation practice of daily care notes are integral to the continuity of patient care because these records provide a comprehensive account of the patient's health trajectory, inclusive of condition status, diagnoses, and treatment plans [1]. Nevertheless, the growing complexity and verbosity of EHR clinical narratives, which are often filled with redundant information, can overwhelm health care providers and increase the risk of diagnostic errors [2-5]. Physicians often skip sections of lengthy and repetitive notes and rely on decisional shortcuts (ie, decisional heuristics) that can contribute to diagnostic errors [6].

Current efforts at automating diagnosis generation from daily progress notes leverage large language models (LLMs). Gao et al [7] introduced a summarization task that takes progress notes as input and generates a summary of active diagnoses. The authors annotated a set of progress notes from the publicly available EHR dataset Medical Information Mart for Intensive Care III (MIMIC-III) [8]. The BioNLP 2023 shared task, known as ProbSum, built upon this work by providing additional annotated notes and attracting multiple efforts focused on developing solutions [9-11]. Demonstrating a growing interest in applying LLMs to serve as solutions, these prior studies use language models such as Text-to-Text Transfer Transformer (T5) [12], developed by Google Research; and Open AI's Generative Pretrained Transformer (GPT) [13]. Unlike the conventional language tasks where LLMs have shown promising abilities, automated diagnosis generation is a critical task that requires high accuracy and reliability to ensure patient safety and improve health care outcomes. Concerns regarding the potential misleading and hallucinated information that could result in life-threatening events prevent LLMs from being used for diagnostic prediction [14].

The Unified Medical Language System (UMLS) [15], a comprehensive resource developed by the National Library of Medicine in the United States, has been extensively used in natural language processing (NLP) research. The UMLS serves as a medical knowledge repository, facilitating the integration, retrieval, and sharing of biomedical information. It offers concept vocabulary and semantic relationships, enabling the construction of medical knowledge graphs (KGs). Prior studies have leveraged UMLS KGs for tasks such as information extraction [16-19] and question answering [17]. Mining relevant knowledge for diagnosis is particularly challenging for 2 reasons: the highly specific factors related to the patient's complaints, histories, and symptoms documented in the EHR; and the vast search space within a KG containing 4.5 million concepts and 15 million relations for diagnosis determination.

In this study, we explore the use of KGs as external resources to enhance LLMs for diagnosis generation. Our work is motivated not only by the potential in the NLP field of augmenting LLMs with KGs [20] but also by the theoretical exploration in medical education and psychology research, shedding light on the diagnostic decision-making process used by clinicians. Forming a diagnostic decision requires the examination of patient data, retrieving encapsulated medical knowledge, and the formulation and testing of the diagnostic hypothesis, which is also known as clinical diagnostic reasoning [21,22]. We propose a novel graph model, DR.KNOWS (Diagnostic Reasoning Knowledge Graph System), designed to retrieve the top N case-specific knowledge paths related to disease pathology and feed them into foundational LLMs to improve the accuracy of diagnostic predictions (as shown in Figure 1). Two distinct foundational models are the subject of this study: T5, known for being fine-tunable; and a sandboxed version of ChatGPT, a powerful LLM where we explore zero-shot prompting.

**Figure 1.** Study overview: we focused on generating diagnoses (text given in red in the "Plan" section) using the SOAP (subjective, objective, assessment, and plan) format progress note with the aid of large language models (LLMs). The input consists of "Subjective," "Objective," and "Assessment" sections (the dotted line box below the heading "Patient Progress Note"), and the diagnoses in the "Plan" section are the ground truth. We introduced an innovative knowledge graph (KG) model, namely DR.KNOWS (Diagnostic Reasoning Knowledge Graph System), that identifies and extracts the most relevant knowledge trajectories from the Unified Medical Language System (UMLS) KG. The nodes of the UMLS KG represent concept unique identifiers (CUIs), and the edges denote the semantic relations among the CUIs. We experimented with prompting ChatGPT for diagnosis generation, with and without the knowledge paths predicted by DR.KNOWS. Furthermore, we investigated how this knowledge grounding influences the diagnostic output of LLMs using human evaluation. The underlined text shows the UMLS concepts identified through a concept extractor. EtOH: ethanol; GI: gastrointestinal; REDCap: Research Electronic Data Capture; T5: Text-to-Text Transfer Transformer; UGIB: upper gastrointestinal bleeding.

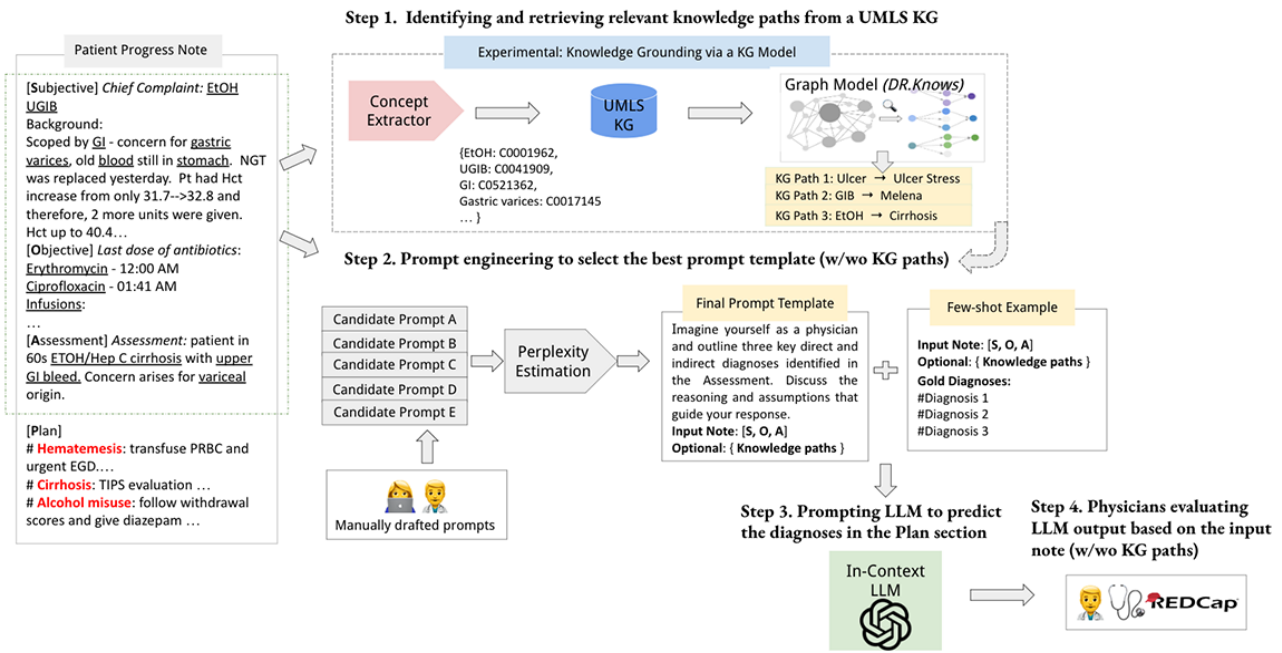

## Objectives

Our work and contribution are structured into two primary components: (1) designing and evaluating DR.KNOWS, a graph-based model that selects the top N probable diagnoses with explainable paths; and (2) demonstrating the usefulness of DR.KNOWS as an additional module to augment pretrained language models in generating relevant diagnoses. Along with the technical contributions, we propose the first human evaluation framework for LLM-generated diagnoses that adapts a survey instrument designed to evaluate diagnostic safety. Our research poses a new exciting problem that has not been addressed in the realm of NLP for diagnosis generation, that is, harnessing the power of KGs for the controllability and explainability of foundational models. By examining the effects of KG path–based prompts on foundational models on a real-world hospital dataset, we strive to contribute to an explainable artificial intelligence (AI) diagnostic pathway.

Several studies have focused on the application of clinical note summarization to discharge summaries [23], hospital course narratives [24], real-time patient visit summaries [25], and problem and diagnosis lists [7,26,27]. Our work follows the line of research on problem and diagnosis summarization. The integration of KGs with LLMs has been gaining traction as an emerging trend due to the potential enhancement of factual knowledge [20], especially on domain-specific question-answering tasks [28-30]. Our work stands out by integrating KGs into LLMs for diagnosis prediction, using a novel graph model for path-based prompts.

## *Methods*

### Problem Formulation

#### *Daily Progress Notes for Diagnosis Prediction*

Daily progress notes are formatted using the SOAP (subjective, objective, assessment, and plan) format [30]. The subjective section of a SOAP daily progress note comprises the patient's self-reported symptoms, concerns, and medical history. The objective section consists of structural data collected by health care providers during observation or examination, such as vital signs (eg, blood pressure and heart rate), laboratory results, or physical examination findings. The assessment section summarizes the patient's overall condition, with a focus on the most active problems and diagnoses for that day. Finally, the plan section contains multiple subsections, each outlining a diagnosis or problem and its treatment plan. Our task is to predict the list of problems and diagnoses that are part of the plan section. Our research used the ProbSum dataset, an annotated resource created for the BioNLP 2023 shared task with gold standard diagnoses derived from progress notes [27].

#### *Using UMLS KGs to Find Potential Diagnoses, Given Medical Narratives*

The UMLS concepts vocabulary comprises >180 sources. For our study, we focused on the Systematized Nomenclature of Medicine–Clinical Terms (SNOMED CT). The UMLS vocabulary is a comprehensive, multilingual health terminology and the US national standard for EHRs and health information exchange. Each UMLS medical concept is assigned a SNOMED

CT concept unique identifier (CUI) from the clinical terminology system. We used semantic types, networks, and semantic relations from UMLS knowledge sources to categorize concepts based on shared attributes, enabling efficient exploration and supporting semantic understanding and knowledge discovery across various medical vocabularies.

Given a medical KG where the nodes represent concepts and the edges denote semantic relations along with an input text describing a patient's problems, we could perform multihop reasoning across the KG and infer the final diagnoses. Figure 2 demonstrates how UMLS semantic relations and concepts can be used to identify potential diagnoses from the evidence provided in a daily care note. The example patient presents with medical conditions of fever, cough and sepsis, which are the concepts recognized by medical concept extractors (Clinical Text Analysis and Knowledge Extraction System [31] and QuickUMLS [32]) and the starting concepts for multihop reasoning. Initially, we extracted the direct neighbors for these concepts. Relevant concepts that aligned with the patient's descriptions were preferred. For precise diagnoses, we chose the top N most relevant nodes at each hop.

**Figure 2.** Problem formulation: inferring possible diagnoses within 2 hops from a Unified Medical Language System (UMLS) knowledge graph given a patient's medical description. The UMLS medical concepts are highlighted in the colored boxes ("female," "sepsis," etc). Each concept has its own subgraph, where concepts are the vertices, and semantic relations are the edges (owing to space constraints, we have omitted the subgraph for "female" in this graph presentation). On the first hop, we could identify the most relevant neighboring concepts to the input description. The darker the color of the vertices, the more relevant they are to the input description. A second hop could be further performed based on the most relevant nodes, leading to the final diagnoses "Pneumonia and influenza" and "Respiratory distress syndrome." Of note, we use the preferred text of concept unique identifiers for presentation purposes. The actual UMLS knowledge graph is built on concept unique identifiers rather than preferred text.

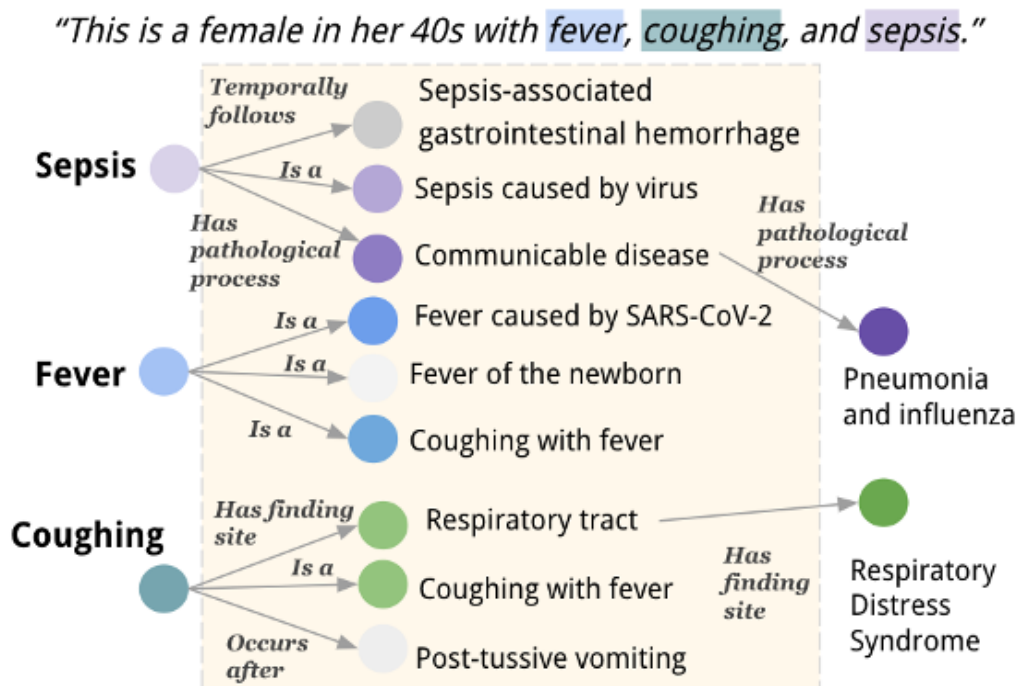

The UMLS's vast repository consists of 270 semantic relations, but not all are crucial for diagnostic reasoning. Adding the nonrelevant relations into a KG introduced substantially complexities in both computation and retrieval processes. A board-certified physician (MA) refined these to identify the 107 most relevant relations for diagnostics, which were then used to build the UMLS KG. This selection, including relations such as "causative agent of" and excluding ones such as "inverse isa," is vital to maintaining computational efficiency and retrieval accuracy within the KG.

## Data Overview

We used 2 sets of progress notes from different clinical settings in this study: MIMIC-III and in-house EHR datasets. MIMIC-III is one of the largest publicly available databases containing deidentified health data from patients admitted to intensive care units. It was developed by the Massachusetts Institute of Technology and Beth Israel Deaconess Medical Center. MIMIC-III includes data from >38,000 patients admitted to intensive care units at the Beth Israel Deaconess Medical Center between 2001 and 2012. The second set, namely the in-house EHR data, was a subset of EHRs that included adult patients (aged 18 years) admitted to the University of Wisconsin health system between 2008 and 2021. In contrast to the MIMIC-III subset, the in-house set covered progress notes from all hospital settings, including the emergency department, general medicine wards, and subspecialty wards. While the 2 datasets originated from separate hospitals and departmental settings and might reflect distinct note-taking practices, both followed the SOAP documentation format for progress notes.

Gao et al [7,9] introduced a subset of 1005 progress notes from MIMIC-III with active diagnoses annotated from the "plan" sections, namely, the ProbSum dataset. Therefore, we applied this dataset for training and evaluation for both graph model intrinsic evaluation and diagnosis summarization. The in-house dataset did not contain human annotation. Even so, by parsing the text with a medical concept extractor that was based on UMLS SNOMED CT vocabulary, we were able to pull out concepts that belonged to the semantic type of "T047 Disease and Syndromes." We deployed this set of concepts as the ground truth data to train and evaluate the graph model. The final in-house dataset contained 4815 progress notes. We present the descriptive statistics in Table 1. When contrasted with MIMIC-III, the in-house dataset exhibited a greater number of CUIs in its input, leading to an extended CUI output. In addition, MIMIC-III encompassed a wider range of abstractive concepts compared to the in-house progress notes.

**Table 1.** Average number of concept unique identifiers (CUIs) in the input and output across the 2 electronic health record datasets: Medical Information Mart for Intensive Care III (MIMIC-III) and in-house. Abstractive concepts are those not found in the input but present in the gold standard diagnoses.

| Datasets | Departments | Input CUIs (n), mean (SD) | Output CUIs (n), mean (SD) | Abstractive CUIs (%) |
| --- | --- | --- | --- | --- |
| MIMIC-III | ICU[a] | 15.95 | 3.51 | 48.92 |
| In-house | All | 41.43 | 5.81 | <1 |

[a]ICU: intensive care unit.

## Graph Model Development

### Overview

This section introduces the architecture design for DR.KNOWS. The DR.KNOWS model is designed to enhance automated diagnostic reasoning by integrating structured clinical knowledge from the UMLS into patient-specific diagnostic predictions. By leveraging a graph-based approach, DR.KNOWS retrieves and ranks relevant knowledge paths from the UMLS, ensuring that only clinically pertinent information is considered. Using a graph neural network, DR.KNOWS incorporates topological information from the UMLS KG into concept representations to better determine each node's relevance to the patient's specific conditions.

### Architecture Overview

As shown in Figure 3, all identified UMLS concepts with an assigned CUI from the input patient text were used to retrieve 1-hop subgraphs from the constructed large UMLS KG. Each node in this graph represents a CUI; therefore, we use "node" and "concept (CUI)" interchangeably throughout. These 1-hop subgraphs are encoded by a stack graph isomorphism network (SGIN) [33], which generates node embeddings that capture both neighboring concept information and pretrained concept embeddings. We chose the SGIN for node embedding because it matches the expressive power of the Weisfeiler-Lehman graph isomorphism test, maximizing the graph neural network's ability to capture meaningful representations. The resulting node embeddings serve as the basis for path embeddings, which the path encoder further processes.

**Figure 3.** DR.KNOWS (Diagnostic Reasoning Knowledge Graph System) model architecture. The input concepts ("female," "fever," etc) are represented by concept unique identifiers (CUIs) represented as a combination of letters and numbers (eg, "C0243026" and "C0015967"). SapBERT: Self-alignment Pretrained Bidirectional Encoder Representations from Transformers.

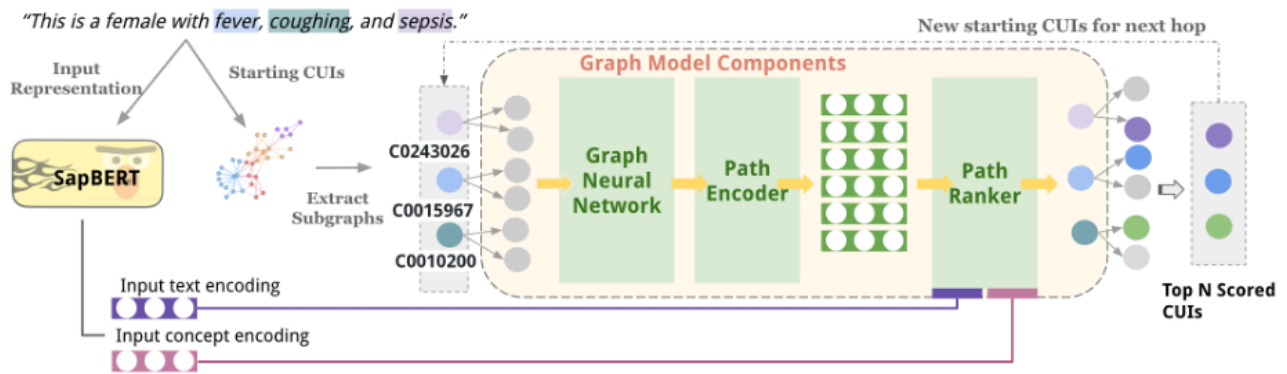

The path encoder module then evaluates these 1-hop paths by examining their semantic and logical alignment with the input text and concept representations, assigning a relevance score to each path. The top N scores across these paths, aggregated across each node's neighboring paths, guide the selection of nodes for the next hop. If no suitable diagnosis node is found, the path exploration terminates by assigning a self-loop to the current node.

While the dominant technique for retrieval-augmented generation systems relies heavily on vector representations and cosine similarity for retrieving and ranking candidate text, our work goes beyond this by adding 2 extra layers of design. First, we leverage the expressive power of the graph structure to enhance the retrieval process. Second, we select paths not simply based on their embeddings but through an attention network that encodes the path-concept relationships, ensuring a more accurate and contextually relevant selection process. In the following paragraphs, we present details regarding each component in the architecture of DR.KNOWS.

### Contextualized Node Representation

We define the deterministic UMLS KG $G = VE$ based on SNOMED CT CUIs and semantic relations, where $V$ is a set of CUIs, and $E$ is a set of semantic relations. Given an input text $x$ containing a set of source CUIs $V_{src} \subseteq V$ and their 1-hop relations $E_{src} \subseteq E$, we can construct relation paths for each source node $v_{src} \subseteq V_{src}$ as $P = \{p_1, p_2,...p_j\}$ such that $p_j = \{v_1, e_1, v_2,...e_{j-1}, v_j\}$, $j \subseteq J$, where $J$ is the maximum length that a source node $v_{src}$ could reach and is nondeterministic. Relations $e$ are encoded as one-hot embeddings. We concatenate all concept names for $v_i$ with special tokens such as [SEP] (for "separator"), such that $l_i$ = [name 1 [SEP] name 2 [SEP]...] and encode $l_i$ using Self-alignment Pretrained Bidirectional Encoder Representations from Transformers (SapBERT) [34] to obtain $h_i$ as concept representation. This allows the CUI representation to serve as the contextualized representation of its corresponding concept names. We chose SapBERT for its contrastive learning-based training, which discriminates similar concepts and their synonyms. It is evaluated on entity linking tasks and has shown state-of-the-art performance. The $h_i$ is further updated through topological representation using the SGIN to become node representation:

$$h_i^{(k)} = MLP^{(k)}\left((1 + \epsilon^{(k)})h_i^{(k)} + \sum_{s \in N(v)} RELU(h_s, e_{s,i})\right)$$

$$h_i = [h_i^{(1)}; h_i^{(2)} ... h_i^{(k)}]$$

$N(v_i)$ represents the set of neighboring nodes of node $v_i$, $h_i^{(k)}$ is the representation of node $v_i$ at layer $k$, $\epsilon^{(k)}$ is a learnable parameter at layer $k$, and $MLP^{(k)}$ is a multilayer perceptron at layer $k$. GIN iteratively aggregates neighborhood information using graph convolution followed by nonlinearity, modeling interactions among nodes within the set $v_{src} \subseteq V_{src}$. Furthermore, the stacking mechanism is introduced to combine multiple GIN layers. The final node representation $v_i$ at layer $K$ (last layer) is computed by stacking the GIN layers, where [...;...] denotes matrix concatenation.

We empirically observed that some types of CUIs are less likely to lead to useful paths for diseases, for example, the concept "recent" (CUI: C0332185) is a temporal concept, and the neighbors associated with it are less useful to predict diagnoses. We designed a weighting scheme based on term frequency–inverse document frequency to assign higher weights to more relevant CUIs and semantic types:

$$W_{CUI} = TFIDF_{concept} * \sum TFIDF_{semantictype_{concept}}$$

$W_{CUI}$ are then multiplied by the corresponding $h_i$ to assign weighted representations to the concept representation.

### Path Reasoning and Ranking

For each node representation $h_i$, we use its n-hop $h_{t,i}^{(n)}$ of the set neighborhood for $v_{t,i}^{(n)}$ for $h_i$ and the associated relation edge $e_{t,i}^{(n)}$ to generate the corresponding path embeddings, with $t$ being the index of the node and its associated neighborhood and relations:

$$p_i = \begin{cases} h_i, & \text{if } n=1 \\ p_{t,i}^{(n-1)}, & \text{otherwise} \end{cases}$$

$$p_{t,i}^{(n-1)} = FFN(W_i h_i^{(n)} + W_t [e_{t,i}^{(n)}, h_{t,i}^{(n)}])$$

where "FFN" is the feedforward network, and $n$ is the number of hops in the subgraph $G_{src}$. The path embedding $p_i$ is the node embedding itself for the first hop and is recursively aggregated with new nodes and edges as the path extends to the next hop.

To determine each path's relevance to the patient's specific symptoms, we used 2 attention mechanisms—multihead attention (MultiAttn) and trilinear attention (TriAttn)—to compute scores $S$ for each path. Both mechanisms use the patient's input text representation $h_x$ and input list of CUIs $h_v$, encoded by SapBERT, to capture explicit and intricate relationships in the input data. MultiAttn was used to explicitly capture relationships between the input text, the list of concepts, and the current path, while TriAttn was used to automatically learn these complex relationships through the inner products of the 3 matrices. As demonstrated in Figure 2, for each hop the path tries to achieve based on the input patient description, the candidate concept can add relevant information, provide no new information and remain neutral, or contradict the information already present in the context.

Using MultiAttn, we define the context relevancy matrix $H_i$ and the concept relevancy matrix $Z_i$ as follows:

$H_i = [h_x; p_i; h_x - p_i; h_x \odot p_i]$

$Z_i = [h_v; p_i; h_v - p_i; h_v \odot p_i]$

$\alpha_i = MultiAttn(H_i \odot Z_i)$,

$S_{Multi} = \varphi(Relu(\sigma(\alpha_i)))$

These relevancy matrices are inspired by a prior work on natural language inference [35], representing logical relations such as neutrality, contradiction, and entailment via matrix concatenation, difference, and product, respectively. Alternatively, TriAttn learns the intricate relations by 3 attention maps:

$\alpha_i = (h_x, h_v, p_i) = \Sigma_{abc} (h_x)_a (h_v)_b (p_i)_c W_{abc}$

$S_{Tri} = \varphi(Relu(\sigma(\alpha_i)))$

$h_x$, $h_v$, and $p_i$ have the same dimensionality D, and $\varphi$ is an MLP player. Finally, we aggregate the MultiAttn or TriAttn scores on all candidate nodes and select the top N nodes (concepts) $V_N$ for the next iteration based on the aggregate attention scores:

$$\beta = Softmax(\Sigma_{i=1}^{V_{src}} \Sigma_{t=1}^{T} S_{i,t})$$

$V_N = argmax_N(\beta)$

By comparing attention scores across candidate paths, the path ranker selects the top N nodes most relevant to each patient's symptoms, maximizing contextual relevance.

*Loss Function*

Our loss function consists of 2 parts: a CUI prediction loss $L_{pred}$ and a contrastive learning loss $L_{CL}$:

$$L = L_{pred} + L_{CL}$$

For CUI prediction loss, we use binary cross entropy loss to calculate whether the predicted node $V_N$ is in the gold standard label $Y$:

$$L_{pred} = \sum_{m}^{M}\sum_{n}^{N}(y_{m,n} * log(v_{m,n}) + (1 - y_{m,n}) * log(1 - v_{m,n}))$$

Where $M$ is the number of sets of gold labels. For contrastive learning loss $L_{CL}$, we encourage the model to learn meaningful and discriminative representations through comparison with positive and negative samples:

$$L_{pred} = \sum_{i} max(\cos(A_i, f_{i+}) - \cos(A_i, f_{i-}) + margin, 0)$$

where $A_i$ is the anchor embedding, defined as $h_x \odot h_v$, representing the input text and concept representation. $\Sigma_i$ indicates a summation over a set of indices $i$, typically representing different training samples or pairs. Inspired by the study by Hu et al [29], we construct $cos(A_i, f_i)$ and $cos(A_i, f_{i-})$ to calculate cosine similarity between $A_i$ and positive feature $f_{i+}$ or negative feature $f_{i-}$, respectively. A positive feature represents the paths correctly leading to the ground truth concept, while a negative feature embodies the paths that, although starting from the source, culminate in an incorrect concept. This equation measures the loss when the similarity between an anchor and its positive feature is not significantly greater than the similarity between the same anchor and a negative feature, considering a margin for desired separation.

We designed a training algorithm to iteratively select and rank the most relevant paths to extend. This algorithm helped to reduce the computational requirement because it does not rank all n-hop paths within 1 pass. This algorithm is presented in Multimedia Appendix 1.

### Selection of Foundational Models and Experiment Setup

Our study centers around the following question: To what extent does the incorporation of DR.KNOWS as a knowledge path–based prompt provider influence the performance of language models in diagnosis summarization?

We present results derived from 2 distinct foundational models, varying significantly in their parameter scales, namely T5-Large, which comprises 770 million parameters [12]; and GPT-3.5-Turbo, which features 154 billion parameters [13]. Specifically, we were granted access to a restricted version of the GPT-3.5-Turbo model, which served as the underlying framework for the highly capable language model, ChatGPT.

These 2 models represent the prevailing direction in the evolution of language models: smaller models such as T5 that offer easier control and larger models such as GPT that generate text with substantial scale and power. Our investigation focused on evaluating the performance of T5 in fine-tuning scenarios and GPT models in zero-shot settings. Our primary objective was not solely to demonstrate cutting-edge results but also to critically examine the potential influence of incorporating predicted paths, generated by graph models, as auxiliary knowledge contributors.

We selected 3 distinct T5-Large variants for fine-tuning using the ProbSum summarization dataset. The chosen T5 models encompass the vanilla T5 [12], a foundational model that has been extensively used in varied NLP tasks; Flan-T5 [36], which has been fine-tuned using an instructional approach; and Clinical-T5 [37], which has been specifically trained on the MIMIC dataset.

Given that our work encompasses a public EHR dataset (MIMIC-III) and a private EHR dataset with protected health information (in-house), we conducted training using 3 distinct computing environments. Specifically, most of the experiments on MIMIC-III were conducted on Google's cloud computing platform, using 1 to 2 NVIDIA A100 40 GB graphics processing units (GPUs) and a conventional server equipped with 1 RTX 3090 Ti 24 GB GPU. The in-house EHR dataset is stored on a workstation located within a hospital research laboratory. The workstation operates within a Health Insurance Portability and Accountability Act–compliant network, ensuring the confidentiality, integrity, and availability of electronic protected health information, and it is equipped with a single NVIDIA V100 32 GB GPU. To use ChatGPT, we used an in-house ChatGPT-3.5-Turbo version hosted on our local cloud infrastructure. No data were sent to Microsoft or OpenAI. This setup ensured that no data were transmitted to OpenAI or external websites, and we were in strict compliance with the MIMIC data use agreement.

While GPT can handle 4096 tokens, T5 is limited to 512 tokens. To ensure a fair comparison, we focused on the subjective and assessment sections of progress notes as input. These sections provide physicians' evaluations of patients' conditions and fall within T5's 512-token limit. This differs from the objective sections, which mainly contain numerical values. Detailed information on data preprocessing, T5 model fine-tuning, and GPT zero-shot setting is presented in Multimedia Appendix 1.

### Prompting Foundational Models to Integrate Graph Knowledge

To incorporate graph model–predicted paths into a prompt, we applied a prompt engineering strategy using domain-independent prompt patterns, as delineated in the study by White et al [38]. Our prompt was constructed with 3 primary components: the output customization that specifies the persona; the output format and template; and the context-control patterns, which are directly linked to the input note and the output of DR.KNOWS. In our test set, for the few input EHRs where no paths could be found (<20 instances), we directly fed the input into the LLMs (T5 and ChatGPT) to generate diagnoses.

Given that our core objective was to assess the extent to which the prompt can bolster the model's performance, it became imperative to test an array of prompts. Gonen et al [39] presented a technique, BETTERPROMPT, which relied on "selecting prompts by estimating language model likelihood." Essentially, we initiated the process with a set of manual task-specific prompts, subsequently expanding the prompt set via automatic paraphrasing facilitated by ChatGPT and backtranslation. We then ranked these prompts by their perplexity score (averaged over a representative sample of task inputs), ultimately selecting those prompts that exhibited the lowest perplexity. Guided by this framework, we manually crafted 5 sets of prompts to integrate the path input, which are visually represented in Table S1 in Multimedia Appendix 1. Specifically, the first 3 prompts were designed by a non–medical domain expert (computer scientist), whereas the final 2 sets of prompts were developed by a medical domain expert (a critical care physician and a medical informaticist). We designated the last 2 prompts (with the medical persona) as "subject matter prompts" and the first 3 prompts as "non–subject matter prompts."

The chosen final prompt came from a template with minimal perplexity, incorporating predicted knowledge paths from the DR.KNOWS model as part of the input. We explored 2 path representation methods: "structural," which uses "→" to link source concepts, edges (relation names), and target concepts; and "clause," which converts paths into clause-style text by directly joining the source and target concepts with their relations. Preliminary experiments showed superior performance with the "structural" representation, leading to its exclusive use in our reported results. The final prompt selected for the foundational models is a paraphrased prompt from the subject matter expert–crafted prompt: "Imagine you are a medical professional equipped with a knowledge graph, and generate the top three direct and indirect diagnoses from the input note. <Input note>…These are knowledge paths: <path 1>; <path 2>…Separate the diagnoses using semicolons, and explain your reasoning starting with <Reasoning>." For the setup where the input did not contain paths, we simply used the prompt with the medical persona and task description as follows: "Imagine you are a medical professional, and generate the top three direct and indirect diagnoses from the input note. <Input note>..." The manually crafted prompts, their paraphrased versions, and their perplexity scores are presented in Table S1 in Multimedia Appendix 1.

## Evaluation Metrics

### Automated Evaluation Metrics for Quantitative Analysis

We conducted 2 evaluations for the DR.KNOWS models: the first was an intrinsic evaluation to determine how many gold standard concepts the graph model can retrieve. The second evaluation examined whether the retrieved knowledge paths could enhance the LLM's diagnosis prediction task. Regarding the first evaluation, our primary objective was to evaluate the effectiveness of DR.KNOWS in predicting diagnoses using CUIs. We used a concept extractor to analyze text within the plan section, specifically extracting CUIs classified under the semantic type T047 DISEASE AND SYNDROMES. We only included CUIs that were guaranteed to connect with at least 1 path, having a maximum length of 2 hops between the target and input CUIs. These chosen CUIs constituted the "gold standard" CUI set, used for both training and assessing the model's performance. As DR.KNOWS predicts the top N CUIs, we measured the Recall@N and Precision@N as follows:

$$Recall = \frac{|\,pred \cap gold\,|}{|\,gold\,|}$$

$$Precision = \frac{|\,pred \cap gold\,|}{|\,pred\,|}$$

The *F*-score, the harmonic mean between recall and precision, will also be reported.

To evaluate foundational model performance on EHR diagnosis prediction, we applied the aforementioned evaluation metric as well as Recall-Oriented Understudy for Gisting Evaluation (ROUGE) [40]. Specifically, ROUGE is a widely used set of metrics designed for evaluating the quality of machine-generated text by comparing it to reference texts. We used the ROUGE–Longest Common Subsequence (ROUGE-L) variant, which is based on the longest common substring; and the ROUGE-2 variant, which focuses on bigram matching. Both ROUGE metrics were used in the ProbSum shared task.

For reporting results from automated metrics, we provided the mean scores across all samples in the test set, along with 95% CIs on 1000 bootstrapped samples.

### Human Evaluation for Qualitative Analysis

Existing evaluation frameworks for AI, such as those used in radiology report generation, do not address diagnosis prediction with LLMs, leaving a significant gap. To address this, our prior work introduced a new human evaluation framework based on the Safer DX Instrument [41], aiming to provide a structured approach for assessing LLMs in diagnosis tasks. In this study, we used this framework to assess the impact of knowledge paths on LLM diagnostic predictions, specifically through a qualitative analysis of the "reasoning" output by LLMs, aiming to gauge the depth and accuracy of the models' diagnostic reasoning processes.

Specifically, we evaluated the model-generated "reasoning" section on the following aspects: (1) *reading comprehension*, (2) *rationale*, (3) *recall of knowledge*, (4) *omission of diagnostic reasoning*, and (5) *abstraction* and *effective abstraction*. *Reading comprehension* was intended to capture whether a model understood the information in a progress note. *Rationale* was intended to capture the inclusion of incorrect reasoning steps. *Recall of knowledge* was intended to capture the hallucination of incorrect facts as well as the inclusion of irrelevant facts in the output. *Omission* of a diagnosis served the same purpose as noted previously by capturing instances when the model failed to support conclusions or provide evidence for a diagnostic choice. *Abstraction* and *effective abstraction* were intended to evaluate the amount of *abstraction* present in each part of the output. This was to ascertain how the knowledge paths influenced the type of output produced and whether the model was able to use abstraction. *Omission* as well as *abstraction* and *effective abstraction* were formatted as *yes* or *no* questions. *Reading comprehension*, *rationale*, and *recall of knowledge* were assessed on a Likert scale ranging from 1 to 5, with 1 indicating strong agreement with poor quality and 5 indicating strong disagreement (representing high quality).

We recruited 2 medical professionals to evaluate LLM outputs using human evaluation guidelines developed by us. Full details of the guidelines, evaluation training, and interannotator agreement are reported in a separate publication (currently under review). The evaluation framework used the REDCap (Research Electronic Data Capture; Vanderbilt University) web application to present the evaluators with input notes, gold standard diagnoses, and model-predicted diagnoses. The evaluators, treated as separate arms in a longitudinal framework, assessed models with KG paths and those without across 2 defined events. Detailed step-by-step guidelines were provided for completing the evaluations in REDCap.

Two senior board-certified clinical informatics physicians served as advisors, pilot testers, and trainers for the 2 medical professionals who completed the human evaluations. The 2 physicians used 5 samples cases to iteratively refine the guidelines provided to the evaluators; these sample evaluations also served as examples for the evaluators to reference during training. The evaluation guidelines consisted of clear descriptions of the meaning of evaluative scores for each aspect of the human evaluation framework as well as a completed example workflow.

## Results

### Intrinsic Evaluation of DR.KNOWS on Predicting Diagnostic Concepts

We compared DR.KNOWS with QuickUMLS, which is a concept extractor baseline that identifies medical concepts from raw text. We took input text, parsed it with QuickUMLS, and outputted a list of concepts. Table 2 presents results from the 2 EHR datasets, MIMIC and in-house. The selection of different top N values was determined by the disparity in text length between the 2 datasets. DR.KNOWS demonstrated superior precision and *F*-scores compared to QuickUMLS across both datasets compared to the baseline, with precision scores of 19.10 (95% CI 17.82-20.37) versus 13.59 (95% CI 12.32-14.88) on the MIMIC dataset and 22.88 (95% CI 20.92-24.85) versus 12.38 (95% CI 11.09-13.66) on the in-house dataset. In addition, its *F*-scores of 25.20 (95% CI 23.93-26.48) on the MIMIC dataset and 25.70 (95% CI 24.06-27.37) on the in-house dataset exceeded the comparison scores of 21.13 (95% CI 19.85-22.41) and 20.09 (95% CI 18.81-21.37), respectively, underscoring the effectiveness of DR.KNOWS in accurately predicting diagnostic CUIs. The TriAttn variant of DR.KNOWS consistently outperformed the MultiAttn variant on both datasets, with *F*-scores of 25.20 (95% CI 23.93-26.48) versus 23.10 (95% CI 21.83-24.39) on the MIMIC dataset and 25.70 (95% CI 24.06-27.37) versus 17.69 (95% CI 16.40-18.96) on the in-house dataset. The concept extractor baseline achieved the highest recall scores—56.91 on the MIMIC dataset and 90.11 on the in-house dataset—because it identified all input concepts that overlapped with the reference CUIs, in particular on the in-house dataset, which was largely an extractive dataset. Training the DR.KNOWS model took an average of 2 of 3 (SD 1.22) hours per epoch on 5000 samples, using 8000 MB of GPU memory.

Table 2. Performance comparison between concept extraction and 2 variants of DR.KNOWS on target concept unique identifier prediction using the Medical Information Mart for Intensive Care (MIMIC-III) and in-house datasets.

| Model | MIMIC-III | | | | In-house | | | |
|---|---|---|---|---|---|---|---|---|
| | Top N knowledge paths | Recall score (95% CI) | Precision score (95% CI) | *F*-score (95% CI) | Top N knowledge paths | Recall score (95% CI) | Precision score (95% CI) | *F*-score (95% CI) |
| Concept extractor | —[a] | 56.91 (55.62-58.18) | 13.59 (12.32-14.88) | 21.13 (19.85-22.41) | — | *90.11*[b] (88.84-91.37) | 12.38 (11.09-13.66) | 20.09 (18.81-21.37) |
| MultiAttn[c] | 4 | 26.91 (25.64-28.19) | *22.79* (21.51-24.06) | 23.10 (21.83-24.39) | 6 | 24.68 (23.35-25.91) | 15.82 (14.55-17.10) | 17.69 (16.40-18.96) |
| MultiAttn | 6 | 29.14 (27.85-30.41) | 16.73 (15.46-18.00) | 19.94 (18.66-21.22) | 8 | 28.69 (27.43-29.98) | 15.82 (14.55-17.11) | 17.33 (16.06-18.60) |
| TriAttn[d] | 4 | 29.85 (26.23-33.45) | 17.61 (16.33-18.89) | 20.93 (19.67-22.21) | 6 | 34.00 (31.04-36.97) | *22.88* (20.92-24.85) | 23.39 (21.71-25.06) |
| TriAttn | 6 | 37.06 (35.80-38.33) | 19.10 (17.82-20.37) | *25.20* (23.93-26.48) | 8 | 44.58 (41.38-47.78) | 22.43 (20.62-24.23) | *25.70* (24.06-27.37) |

[a]Not applicable.

[b]Best performance values are italicized.

[c]MultiAttn: multihead attention.

[d]TriAttn: trilinear attention.

## Assessing the Impact of DR.KNOWS on Diagnosis Prediction

The best systems for each foundational model on the ProbSum test set are presented in Table 3, including those with predicted paths provided by DR.KNOWS and those without. Overall, the prompt-based fine-tuning of T5 surpassed ChatGPT's prompt-based zero-shot approach on all metrics, and ChatGPT's prompt-based few-shot approach showed comparable performance to T5. Notably, models that incorporated paths, particularly for the CUI *F*-score, showed significant improvements. The vanilla T5 model with a path prompt excelled, achieving the highest ROUGE-L score (30.72, 95% CI 30.40-32.44) and CUI *F*-score (27.78, 95% CI 27.09-29.80). This ROUGE-L score could have ranked third on the ProbSum leaderboard [27], which is noteworthy considering that the top 2 systems used ensemble methods [10,11].

**Table 3.** Best performance on the Medical Information Mart for Intensive Care III (MIMIC III) test set (with annotated active diagnoses) from 3 Text-to-Text Transfer Transformer (T5) variants and ChatGPT across all prompt styles with DR.KNOWS (Diagnostic Reasoning Knowledge Graph System) path prompting and without. To illustrate the performance differences better, we report Recall-Oriented Understudy for Gisting Evaluation-2 (ROUGE-2); ROUGE–Longest Common Subsequence (ROUGE-L); and concept unique identifier (CUI) recall, precision, and F-scores.

| Model | Rouge-2 score (95% CI) | Rouge-L score (95% CI) | CUI recall score (95% CI) | CUI precision score (95% CI) | CUI *F*-score (95% CI) |
| --- | --- | --- | --- | --- | --- |
| **Prompt-based fine-tuning setting** | | | | | |
| Vanilla T5 | 12.66 (11.24-13.54) | 29.08 (27.52-29.99) | 39.17 (37.53-41.56) | 22.89 (21.02-23.62) | 26.19 (25.31-26.78) |
| Vanilla T5+path[a] | 13.13 (12.64-13.88) | *30.72*[b] (30.40-32.44[c]) | *40.73* (39.46-42.18) | 24.28 (23.49-26.03) | *27.78* (27.08-29.80[c]) |
| Flan-T5 | 11.83 (10.51-12.40) | 27.02 (25.64-27.80) | 38.28 (36.70-39.45) | 22.32 (21.81-23.00) | 25.32 (24.10-26.34) |
| Flan-T5+path | *13.30* (12.19-14.44) | 30.00 (29.20-32.70[c]) | 38.96 (37.48-40.01) | *24.74* (23.35-26.12[c]) | 27.38 (26.98-28.68[c]) |
| Clinical-T5 | 11.68 (11.06-12.49) | 25.84 (23.74-26.15) | 30.37 (28.94-30.99) | 17.91 (15.46-19.79) | 19.61 (16.44-20.03) |
| Clinical-T5+path | 12.06 (10.89-12.48) | 25.97 (24.71-26.33) | 29.45 (27.65-30.19) | 22.78 (21.35-23.59[c]) | 23.17 (21.39-23.96[c]) |
| **Prompt-based zero-shot setting** | | | | | |
| ChatGPT | 7.05 (6.54-7.56) | 19.77 (19.26-20.28) | 23.68 (23.18-24.19) | 15.52 (15.00-16.02) | 16.04 (15.53-16.55) |
| ChatGPT+path | 5.70 (5.19-6.21) | 15.49 (14.98-15.99) | 25.33 (24.82-25.84[c]) | 17.05 (16.29-17.81[c]) | 18.21 (17.46-18.98[c]) |
| **Prompt-based few-shot setting** | | | | | |
| ChatGPT 3-shot | 9.63 (8.32-10.06) | 21.84 (19.99-22.09) | 22.71 (20.99-23.96) | 19.57 (17.23-19.78) | 21.02 (20.26-21.79) |
| ChatGPT 5-shot | 9.73 (8.52-10.18) | 21.23 (19.58-21.72) | 22.45 (20.93-23.80) | 19.67 (17.66-20.33) | 20.96 (20.19-21.73) |
| ChatGPT 3-shot+path | 10.66 (9.17-10.72) | 24.32 (22.44-24.25[c]) | 26.48 (25.33-28.36[c]) | 24.22 (21.44-24.21[c]) | 25.30 (24.52-26.06[c]) |
| ChatGPT 5-shot+path | 11.73 (10.51-12.25[c]) | 25.43 (23.53-25.35[c]) | 27.76 (26.56-29.39[c]) | 24.56 (22.47-25.12[c]) | 26.02 (25.25-26.78[c]) |

[a]Prompt styles with DR.KNOWS path prompting.

[b]Best performance values are italicized.

[c]95% CIs with a distinct CI for the DR.KNOWS-prompted path compared to no-path scenarios.

The comparison between ChatGPT with DR.KNOWS and ChatGPT without in the predicted paths scenario provided additional insights. In the few-shot setting, the incorporation of paths led to marked improvements; for instance, in the 3-shot setting, the with-path scenario outperformed the no-path scenario on all metrics, with ROUGE-L score of 24.32 (95% CI 22.44-24.25) compared to ChatGPT 3-shot no-path ROUGE-L score of 21.84 (95% CI 19.44-22.09) and CUI *F*-score of 25.30 (95% CI 24.52-26.06) versus 21.02 (95% CI 20.26-21.79). In the 5-shot setting, ChatGPT with paths achieved a ROUGE-L score of 25.43 (95% CI 25.53-25.35) compared to 21.23 (95% CI 19.58-21.72) for ChatGPT without paths and CUI *F*-score of 26.02 (95% CI 25.25-26.78) versus 20.96 (95% CI 20.19-21.73).

## Human Evaluation Results

After the annotation procedure, the 2 medical professionals completed a supervised set of evaluations and were considered validated once they achieved a κ coefficient of 0.7 with the physician trainers and each other.

Although the T5 and ChatGPT models displayed similar performance on automated metrics, their outputs diverged significantly. The T5 models, lacking instruction tuning, failed to respond adequately to prompts requesting the generation of a <Reasoning> section. Consequently, our human evaluation focused exclusively on the outputs produced by ChatGPT. We conducted human evaluation of the top-performing ChatGPT output (5-shot approach), comparing scenarios with the DR.KNOWS knowledge paths with KG and without KG. The final evaluation set consisted of 92 input notes and 2 sets of ChatGPT-predicted text.

The results are reported in Figure 4. First, there was no significant increase in *omission of diagnoses*, with 16% (15/92) observed with KG as opposed to 10% (9/92) without KG (*P*=.16). Regarding *rationale* (correct reasoning), ChatGPT with KG exhibited stronger agreement with the human evaluators (51/92, 55%) than ChatGPT without KG (46/92, 50%; *P*<.001). In the *abstraction* category (assessing the presence of abstraction in the model output), there was a notable drop from 88% (81/92; without KG ) to 78% (71/92; with KG ) in the affirmative responses (*P*=.03), indicating that less abstraction was required when KG paths were included. Differences were also noted in *effective abstraction* in favor of the KG paths (*P*=.002).

**Figure 4.** Human evaluation of ChatGPT outputs comparing scenarios with ("KG" [knowledge graph]) the DR.KNOWS (Diagnostic Reasoning Knowledge Graph System) knowledge paths and without ("No KG").

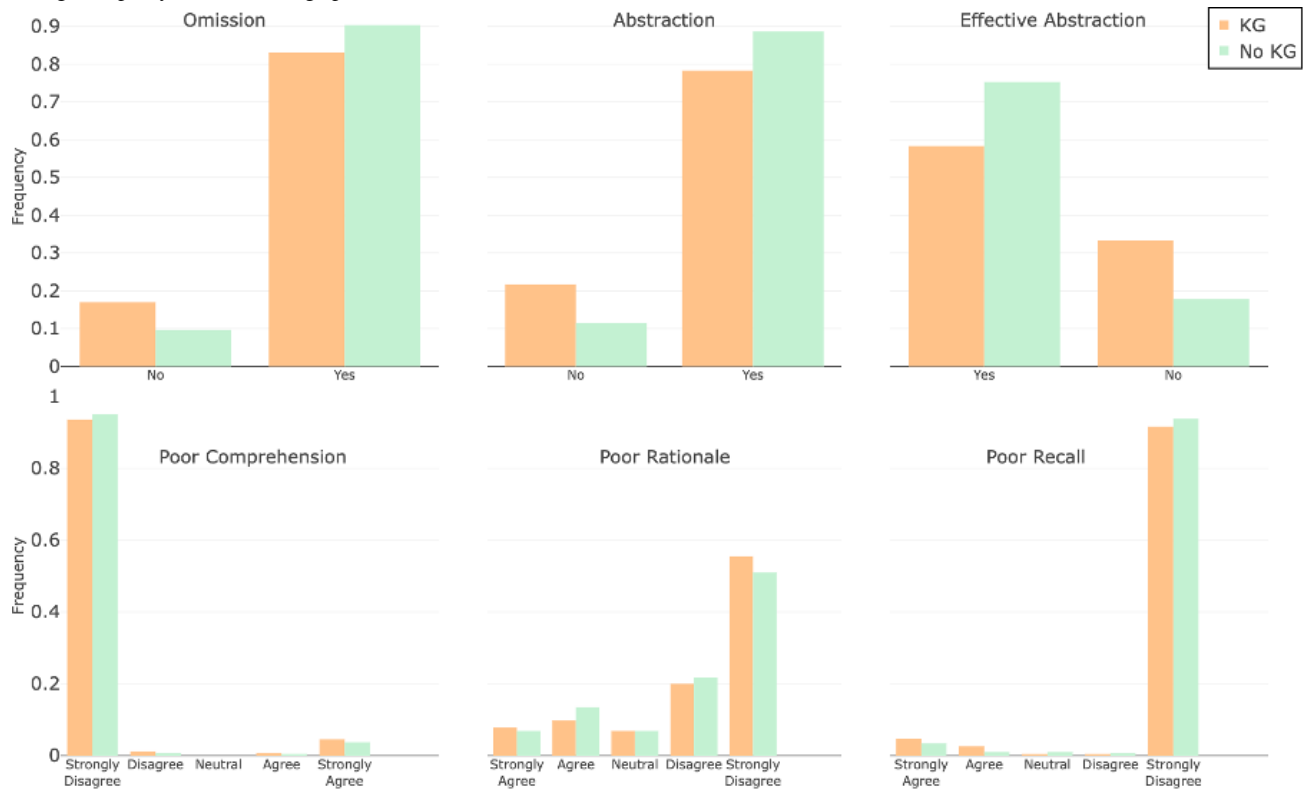

### Error Analysis

We discovered 2 primary types of errors in the DR.KNOWS outputs that could result in missed opportunities for improving knowledge grounding. Figure 5 presents an example where ChatGPT did not find the provided knowledge paths useful. In this case, the majority of the provided knowledge paths were highly extractive ("leukocytosis," "reticular dysgenesis," and "paraplegia" are the target concepts to which the knowledge paths led, and all are associated with a "self-loop" relationship). On the abstraction paths, the retrieved target concepts "abdomen hernia scrotal" and "chronic neutrophilia" were not relevant to the input patient condition.

**Figure 5.** An example of an error in the knowledge paths retrieved by DR.KNOWS (Diagnostic Reasoning Knowledge Graph System). DR.KNOWS retrieved 2 paths leading to irrelevant and misleading diagnoses (marked in red). The counterclockwise gapped circular arrow symbol represents a self-loop.

**Input progress note:**
<Assessment> 73 yo M w/ mmp, C4-5 paraplegia, TF dependence, on broad spectrum abx for recent pna transferred from OSH w/ resp distress, leukocytosis, and HOTN <Subjective> Chief Complaint: resp distress and hypotension I saw and examined the patient, and was physically present with the ICU Resident for key portions of the services provided. I agree with his / her note above, including assessment and plan. HPI: 73 yo M w/mmp including C4-5 paraplegia, TF dependence, on broad spectrum abx for recent pna transferred from OSH w/ resp distress, leukocytosis, and HOTN. 24 Hour Events: PICC LINE - START 04:24 PM ARTERIAL LINE - START 07:00 PM HOTN responded to IVF, never required pressor support MS ANTIBX coverage broadened--> vanco/ CT chest ordered Leukocytosis normalizing History obtained from Medical records Allergies: Methyldopa hives; Shellfish pt. with remote.

**Dr.Knows retrieved top-6 knowledge paths:**

Leukocytosis → self → Leukocytosis ↺ <path>

reticular dysgenesis → self → reticular dysgenesis ↺ <path>

Leukocytosis → definitional manifestation of → Leukocytosis ↺ <path>

Paraplegia → self → Paraplegia ↺ <path>

Thoracic → has finding site → abdomen hernia scrotal ↺ <path>

Leukocytosis → definitional manifestation of → Leukocytosis → has definitional manifestation → Chronic neutrophilia

Another error observed occurred when DR.KNOWS selected the source CUIs that were less likely to generate pertinent paths for clinical diagnoses, resulting in ineffective knowledge paths. Figure 6 shows a retrieved path from "consulting with (procedure)" to "consultation-action (qualifier value)." Although some procedure-related concepts such as endoscopy or blood testing were valuable for clinical diagnosis, this specific path of consulting did not contribute meaningfully to the input case.

Similarly, another erroneous pathway began with "drug allergy" and led to "allergy to dimetindene (finding)," which is contradictory, given that the input note explicitly states "no known drug allergies." While the consulting path's issue was its lack of utility, the "drug allergy" path could introduce the risk of hallucination (misleading or fabricated content) within ChatGPT.

**Figure 6.** An example illustrating ChatGPT's performance with the knowledge paths extracted by DR.KNOWS (Diagnostic Reasoning Knowledge Graph System). Two paths had source concept unique identifiers ("Consulting with [procedure]" and "Drug allergy") that were less likely to generate pertinent paths for clinical diagnoses. Of note, the path of "Drug allergy" led to a path contradicting the "No Known Drug Allergies" description in the input. The path of "cirrhosis of liver" represents a correct diagnosis, but ChatGPT failed to include it. The counterclockwise gapped circular arrow symbol represents a self-loop. ESRD: end-stage renal disease.

```
Input progress note:
<Assessment> 57M with Hep C cirrhosis, ESRD on HD, presenting with hypotension and shock, elevated lactate, and drop in hematocrit.
<Subjective> TITLE: Chief Complaint: Hypotension 24 Hour Events: - Levophed not able to be weaned - PT consult - Ordered VBG with O2 sat and
lactate to evaluate whether he's ischemic during all this hypoTN  Allergies: No Known Drug Allergies

Dr.Knows retrieved top-6 knowledge paths:
Unspecified chronic renal failure → possibly equivalent to → Renal failure: [chronic] or [end stage] → possibly equivalent to → Unspecified
chronic renal failure <path>

Cirrhosis of liver (disorder) → self → Cirrhosis of liver (disorder) ↺ <path>
Allergic reaction (disorder) → self → Allergic reaction (disorder) ↺ <path>
Unspecified chronic renal failure → possibly equivalent to → Renal failure: [chronic] or [end stage] ↺ <path>
Consulting with (procedure) → method of → Consultation - action (qualifier value) ↺ <path>
Drug allergy → has definitional manifestation → Allergy to dimetindene (finding) ↺

Gold Standard Diagnosis:
Hypotension/shock. Most Likely septic shock; ESRD; Cirrhosis

Predicted Diagnoses (with knowledge paths input):
ESRD with hypotension and shock; elevated lactate; drop in hematocrit.
```

In addition to the errors in the DR.KNOWS outputs, there were instances where ChatGPT failed to leverage the accurate knowledge paths presented. Figure 6 includes a knowledge path regarding "cirrhosis of liver," which was the correct diagnosis. However, ChatGPT response did not include this diagnosis.

## Discussion

### Principal Findings

DR.KNOWS showed significant advantages over the QuickUMLS concept extractor baseline in extracting correct concepts for diagnoses. On the ProbSum dataset, where the goal was to generate a list of diagnoses given the progress notes, prompt-based fine-tuning of T5 outperformed ChatGPT's zero-shot approach and showed comparable results to its few-shot approaches, with the inclusion of predicted paths by DR.KNOWS significantly enhancing performance across all metrics. The vanilla T5 with path prompts notably achieved top ROUGE-L and CUI $F$-scores, demonstrating the effectiveness of incorporating paths into the model. Human evaluation of ChatGPT's reasoning section showed strong agreement with human evaluators in terms of correct *rationale* and enhanced *effective abstraction*, indicating nuanced improvement in reasoning and abstraction quality with KG integration.

While DR.KNOWS leverages KG paths to enhance diagnosis prediction, it is important to acknowledge the potential biases and limitations inherent in KG data. KGs such as UMLS are comprehensive, but they may reflect biases based on the clinical domains and patient populations from which they were constructed, which could impact the relevance or appropriateness of the retrieved paths. To mitigate this, DR.KNOWS focuses on case-specific path selection, aiming to retrieve only the paths most directly relevant to the patient context. Nonetheless, future iterations could benefit from evaluating path relevance using additional contextual information, such as demographic details, to better align with patient-specific needs and reduce bias.

Error analysis showed that DR.KNOWS occasionally struggled with identifying knowledge paths unrelated to the patient representation; in addition, the analysis emphasized the importance of selecting accurate starting medical concepts. Currently, DR.KNOWS relies solely on semantic-based ranking on the candidate paths, that is, the cosine similarity between candidate path embeddings and input text, with the embedding quality being crucial for ranking performance. Improving the representation and embedding methods, as well as exploring probabilistic modeling techniques [42,43], could enhance path relevance. Furthermore, incorporating a graph reasoning mechanism that enables symbolic chain-of-thought reasoning might compensate for the weaknesses of contextualized embeddings and cosine-similarity metrics [44], presenting a valuable future direction. This integration could improve the diagnostic potential of DR.KNOWS, allowing for more nuanced and bias-aware reasoning.

The error analysis also presented instances where ChatGPT neglected to incorporate certain beneficial knowledge paths. It is important to acknowledge that ChatGPT operates as a black box application programming interface model, with its internal weights and training processes being inaccessible. To enhance the efficacy of the graph-based retrieve-and-augment framework, it would be advantageous to explore the potential of graph prompting and instruction tuning on open-source language models. These methods could refine the model's ability to use relevant information effectively. Other relevant research also uses advanced prompting techniques, such as self-retrieval–augmented generation [45] and step-back prompting [46]. The Google Research team recently presented a study investigating multiple ways of encoding graphs into LLM inputs [47], which might inform a future direction for this work beyond the typical structural or clause-based path prompting.

In conclusion, LLMs such as ChatGPT hold promise for generating diagnoses for clinical decision support; however, methods such as graph prompting are needed to guide the model down the correct reasoning paths to avoid hallucinations and provide comprehensive diagnoses. While we show some progress in a graph prompting approach with DR.KNOWS, more work is needed to improve methods that leverage the UMLS knowledge source for grounding to achieve more accurate outputs. Nonetheless, DR.KNOWS represents a step toward trustworthy AI in medicine, providing knowledge grounding to LLMs and potentially reducing factual errors in diagnostic outputs [48]. Furthermore, our proposed human evaluation framework, derived from diagnostic safety evaluations used in clinical settings, enables the assessment of LLMs from the perspective of diagnostic safety. It carries strong face validity and reliability to evaluate a model's strengths and weaknesses as a diagnostic decision support system. This ensures that the models not only perform well on technical metrics but also align with clinical standards of safety and reliability.

## Limitations

Our work on leveraging KGs for LLM diagnosis generation has shown promising results; however, there are notable limitations that must be acknowledged. First, while the UMLS concept extractors (Clinical Text Analysis and Knowledge Extraction System and QuickUMLS) are powerful tools, they are not without flaws. One significant limitation is their inability to accurately identify all relevant concepts, particularly indirect or nuanced medical concepts. These indirect concepts can be crucial for accurate diagnosis generation; yet, the current concept extractors may fail to recognize them, leading to incomplete or less accurate knowledge representation.

Second, our path selection methodology relies heavily on cosine similarity, a common approach within the retrieval-augmented generation framework. Despite its prevalence, this method has inherent limitations due to its heavy reliance on the quality of embedding representations. If the embeddings do not adequately capture the semantic nuances of medical concepts, the similarity measure may lead to the retrieval of less relevant or noisy knowledge paths. This can ultimately impact the quality and reliability of the diagnostic suggestions generated by the LLM.

These limitations highlight the need for the continued refinement of both the concept extraction and path selection processes. Future work should explore more sophisticated techniques to enhance concept identification and improve the robustness of embedding representations, thereby reducing the reliance on cosine similarity and increasing the overall accuracy and utility of the KG-based approach.

## Acknowledgments

This work is supported by grants from the National Institutes of Health. Funding was supported by the National Library of Medicine (K99LM014308, R00LM014308: YG; R01LM012973-04: TM and DD); the National Heart, Lung, and Blood Institute (R01HL157262-03: MMC); and the National Institute on Drug Abuse (R01DA051464: MA).

## Data Availability

The source code knowledge graph generated during this study are available on the GitHub repository [49]. Medical Information Mart for Intensive Care III is available from PhysioNet.

## Authors' Contributions

YG was responsible for conceptualization, supervision, methodology, formal analysis, writing (original draft as well as review and editing), validation, visualization, data curation, investigation, project administration, and funding acquisition. RL was responsible for writing (original draft as well as review and editing), methodology, data curation, validation, investigation, conceptualization, and formal analysis. EC was responsible for writing (original draft as well as review and editing), validation, methodology, data curation, investigation, conceptualization, and formal analysis. JRC was responsible for writing (review and editing), formal analysis, investigation, and data curation. BWP was responsible for writing (review and editing), validation, formal analysis, methodology, investigation, and conceptualization. MMC was responsible for writing (review and editing), conceptualization, methodology, and funding acquisition. TM was responsible for writing (review and editing), conceptualization, methodology, and funding acquisition. DD was responsible for writing (review and editing), conceptualization, methodology, and funding acquisition. MA was responsible for conceptualization, supervision, methodology, formal analysis, writing (original draft as well as review and editing), validation, visualization, data curation, investigation, project administration, and funding acquisition.

**Conflicts of Interest**

TM is a consultant for Lavita.ai, a startup that builds NLP tools for medical use cases. All other authors declare no conflicts of interest.

**Multimedia Appendix 1**

Data preprocessing, DR.KNOWS (Diagnostic Reasoning Knowledge Graph System) training details, prompt engineering using ChatGPT, and Text-to-Text Transfer Transformer (T5) fine-tuning.
[[DOCX File , 37 KB-Multimedia Appendix 1](#)]

## Abbreviations

**AI:** artificial intelligence
**CUI:** concept unique identifier
**DR.KNOWS:** Diagnostic Reasoning Knowledge Graph System
**EHR:** electronic health record
**GPT:** Generative Pretrained Transformer
**GPU:** graphics processing unit
**KG:** knowledge graph
**LLM:** large language model
**MIMIC-III:** Medical Information Mart for Intensive Care III
**MultiAttn:** multihead attention
**REDCap:** Research Electronic Data Capture
**ROUGE:** Recall-Oriented Understudy for Gisting Evaluation
**ROUGE-L:** Recall-Oriented Understudy for Gisting Evaluation–Longest Common Subsequence
**SapBERT:** Self-alignment Pretrained Bidirectional Encoder Representations from Transformers
**SGIN:** stack graph isomorphism network
**SNOMED CT:** Systematized Nomenclature of Medicine–Clinical Terms
**SOAP:** subjective, objective, assessment, and plan
**T5:** Text-to-Text Transfer Transformer
**TriAttn:** trilinear attention
**UMLS:** Unified Medical Language System